\documentclass[runningheads]{llncs}
\usepackage{eccv}

\usepackage{eccvabbrv}
\usepackage{glossaries}
\usepackage{graphicx}
\usepackage{booktabs}
\usepackage{comment}

\usepackage[accsupp]{axessibility}  % Improves PDF readability for those with disabilities.

\usepackage[pagebackref,breaklinks,colorlinks,citecolor=eccvblue]{hyperref}
\usepackage{orcidlink}

\usepackage{amsmath}
\usepackage{array}
\usepackage{multirow}

\newacronym{crp}{CRP}{Concept Relevance Propagation}
\newacronym{pcx}{PCX}{Prototypical Concept-based eXplanations}
\newacronym{lrp}{LRP}{Layer-wise Relevance Propagation}
\newacronym{lpips}{LPIPS}{Learned Perceptual Image Patch Similarity}

\begin{document}

% ---------------------------------------------------------------
\title{{Synthetic Generation of Dermatoscopic Images with GAN and Closed-Form Factorization}}

\titlerunning{Synthetic Generation of Dermatoscopic Images}

\author{
Rohan Reddy Mekala \inst{1}\orcidlink{0000-0002-2274-9560} \and % rreddy@fraunhofer.org
Frederik Pahde \inst{2}\orcidlink{0000-0002-5681-6231} \and % frederik.pahde@hhi.fraunhofer.de
Simon Baur \inst{2}\orcidlink{0009-0009-4307-3078}\and % simon.baur@hhi.fraunhofer.de
Sneha Chandrashekar \inst{1}\orcidlink{0009-0001-7835-6061} \and %  https://orcid.org/0009-0001-7835-6061 ? Schandrashekar.external-partner@fraunhofer.org
% Others alphabetically:
Madeline Diep \inst{1}\orcidlink{0000-0002-9908-0367} \and % https://orcid.org/0000-0002-9908-0367 mdiep@fraunhofer.org
Markus Wenzel \inst{2}\orcidlink{0000-0002-6540-1476} \and % markus.wenzel@hhi.fraunhofer.de
Eric L. Wisotzky \inst{2} \orcidlink{0000-0001-5731-7058} \and % eric.wisotzky@hhi.fraunhofer.de
Galip Ümit Yolcu \inst{2} \orcidlink{0009-0004-1358-8955} \and % galip.uemit.yolcu@hhi.fraunhofer.de
% Group leaders alphabetically:
Sebastian Lapuschkin \inst{2}\orcidlink{0000-0002-0762-7258} \and % sebastian.lapuschkin@hhi.fraunhofer.de
Jackie Ma \inst{2}\orcidlink{0000-0002-2268-1690} \and % jackie.ma@hhi.fraunhofer.de
% Heads alphabetically:
Peter Eisert \inst{2}\orcidlink{0000-0001-8378-4805} \and % peter.eisert@hhi.fraunhofer.de
Mikael Lindvall \inst{1}\orcidlink{0009-0001-7457-4002} \and % % https://orcid.org/0009-0001-7457-4002 mlindvall@fraunhofer.org
Adam Porter \inst{1} \and % \orcidlink{} aporter@fraunhofer.org
Wojciech Samek \inst{2}\orcidlink{0000-0002-6283-3265} % wojciech.samek@hhi.fraunhofer.de
}

\authorrunning{R.~R.~Mekala et al.}

\institute{Fraunhofer USA Center Mid-Atlantic, 20737-1250 Riverdale, MD, USA\\
\email{rreddy@fraunhofer.org}\\
\url{https://www.cma.fraunhofer.org/} \and
Fraunhofer Heinrich-Hertz-Institut, 10587 Berlin, Germany\\
\email{markus.wenzel@hhi.fraunhofer.de}\\
\url{https://www.hhi.fraunhofer.de/}}

\maketitle

% Abstract is at most 4000 characters
% Papers are limited to 14 pages, including figures and tables (not including references), in the LNCS style of Springer.

\begin{abstract}
  In the realm of dermatological diagnoses, where the analysis of dermatoscopic and microscopic skin lesion images is pivotal for the accurate and early detection of various medical conditions, the costs associated with creating diverse and high-quality annotated datasets have hampered the accuracy and generalizability of machine learning models. We propose an innovative unsupervised augmentation solution that harnesses Generative Adversarial Network (GAN) based models and associated techniques over their latent space to generate controlled ``semi-automatically-discovered'' semantic variations in dermatoscopic images. We created synthetic images to incorporate the semantic variations and augmented the training data with these images. With this approach, we were able to increase the performance of machine learning models and set a new benchmark amongst non-ensemble based models in skin lesion classification on the HAM10000 dataset; and used the observed analytics and generated models for detailed studies on model explainability, affirming the effectiveness of our solution.
\end{abstract}

\section{Introduction}
\label{sec:intro}

The application of artificial intelligence (AI) and machine learning (ML) in the medical domain has garnered substantial interest due to its potential to aid health practitioners in diagnosing conditions, predicting patient outcomes, and personalizing patient care. In dermatology, AI and ML have demonstrated superior performance compared to dermatologists in analyzing dermatoscopy images \cite{brinker2019deep}.
AI/ML models can process vast quantities of images rapidly, assisting dermatologists in making faster and more accurate diagnoses, thereby improving patient care, and potentially saving lives. However, the success of such AI/ML models is fundamentally limited by the lack of availability of datasets with sufficient variations to reflect semantic occurrences in the real world. Additionally, privacy concerns and regulatory constraints pose a major hindrance towards procuring additional annotated medical image datasets, making it necessary to explore alternate ways of synthesizing these image variants in a reliable manner, while ensuring the photorealism and fidelity of the generated images.

In the domain of medical imaging, the concept of synthetic data generation has manifested remarkable strides across various disciplines and applications \cite{yi2019generative, jeong2022systematic, chen2022generative, solanki2023gans, makhlouf2023use, kebaili2023deep}. Image synthesis, particularly through methods such as GANs and diffusion models, makes it possible to augment low-volume training data. These generative approaches have also been explored within the realms of dermatoscopy \cite{yi2018unsupervised, baur2018melanogans, baur2018generating, abhishek2019mask2lesion, qasim2020red-gan, sagers2022improving, akrout2023diffusion} and histopathology diagnostics \cite{levine2020synthesis, quiros2021pathologyGAN, dolezal2023deep, aversa2023diffinfinite}.

However, existing methods for developing transformations in the GAN latent space predominantly rely on classification models to ensure the generated images have specific attributes. While these models have been instrumental in driving advances in image synthesis and manipulation, they come with significant drawbacks. Classification-based methods require large amounts of labeled data, which are often difficult to obtain due to privacy concerns, regulatory constraints, and the high costs associated with manual annotation. This reliance on labeled data can severely limit the scalability and applicability of these models, particularly for medical data modalities like dermatoscopy where annotated datasets for various style variations are scarce. Moreover, since classification models are constrained to predefined categories of semantics, the scope of transformations that can be learned is considerably restricted. This results in dependence on domain experts to identify semantics, with the added costs of procuring and annotating images with such semantics.

In this paper, we present a novel approach towards developing variations in medical images using this unsupervised method based on the latent space of GANs. Our approach leverages the capabilities of two advanced GAN models: StyleGAN2 \cite{karras2020analyzing} and HyperStyle \cite{alaluf2022hyperstyle}. Initially, we train the StyleGAN2 model on a comprehensive dataset of dermatoscopic images to generate high-quality synthetic images. Following this, we employ HyperStyle for GAN inversion, optimizing latent features extracted from real images. We then implement closed-form factorization to identify meaningful and orthogonal latent semantic directions within the latent space. Finally, we validate and refine these directions to ensure they correspond to human-understandable and domain-relevant transformations. Our research extends beyond the realm of image generation, addressing the crucial need for evaluation metrics in the context of synthetic skin lesion images. We assess the perceptual similarity of the generated images using state-of-the-art metrics such as the \gls{lpips}.
These metrics provide a quantitative foundation for evaluating the fidelity of synthetic images in comparison to their real counterparts.

To further show the efficacy of our approach, we train classification models on the augmented dataset, achieving state-of-the-art performance in lesion classification. This pipeline not only mitigates the challenges associated with traditional classification models but can potentially enhance the scalability, efficiency, and interpretability of transformation development in medical image analysis.
Thus, this research work has three important contributions:
\begin{itemize}
\item Through generating high-fidelity synthetic skin lesion images, we pioneer the application of advanced GAN models \cite{karras2020analyzing, alaluf2022hyperstyle}, and explore the effectiveness of these models in capturing nuanced details and variations in skin lesions.
\item By identifying transformations relevant to the skin lesion domain, we contribute to the field of unsupervised transformation development. These transformations are crucial for data augmentation and enhancing the diversity of synthetic skin lesion images.
\item We demonstrate the practical impact of synthetic data in improving the performance and explainability of machine learning models for skin lesion analysis. The transformed images significantly contributed to the training of a skin lesion classification model, resulting in a notable increase in accuracy compared to conventional datasets.
\end{itemize}

In the following sections, we describe our method in greater detail, apply it to a case study, report on the result, and discuss potential future work.

\section{Our Approach}

\subsection{Background}
GANs have revolutionized the field of medical imaging by providing innovative augmentation-driven solutions to data scarcity and enhancing the quality of synthetic medical images. These GAN-based solutions have had a particular impact in domains such as radiology, pathology, and dermatology, where obtaining high-quality labeled data is often challenging due to privacy concerns, regulatory constraints, and the high cost of manual annotation \cite{yi2019generative}. In dermatoscopy, GANs have shown significant promise in synthesizing skin lesion images, which can be used to augment existing datasets and improve the performance of diagnostic models \cite{tschandl2018ham10000, baur2018melanogans}.

A typical GAN architecture \cite{goodfellow2014generative} consists of two neural networks: the generator and the discriminator which work collaboratively through an adversarial training process. The generator creates new data samples, while the discriminator evaluates them against real data to train the generator to produce images aimed at being indistinguishable from real images. This adversarial process continues until the generator produces high-quality realistic images.

Our approach leverages two state-of-the-art GAN models: StyleGAN2 \cite{karras2020analyzing} and HyperStyle \cite{alaluf2022hyperstyle}. StyleGAN2 stands out due to its architectural innovations, which include redesigned generator normalization, progressive growing, and the introduction of a style-based generator architecture. These enhancements enable the generation of highly realistic and detailed images by allowing the model to control different levels of detail through the latent space. Compared to Variational Autoencoders (VAEs), StyleGAN2 generally produces higher quality images with sharper details and more coherent structures. While VAEs are effective for generating diverse samples, they often suffer from blurrier outputs. On the other hand, compared to conditional diffusion models, StyleGAN2 typically achieves faster generation times and requires less computational resources, as diffusion models often involve iterative processes that are more computationally intensive. These advantages make StyleGAN2 particularly suitable for applications requiring high fidelity and variability of the generated images. On the other hand, HyperStyle focuses on the challenge of image inversion, which involves mapping real images into the latent space of a GAN which is used by the generator to manipulate the image. HyperStyle employs a hybrid approach that combines the strengths of encoder- and optimization-based inversion techniques. By balancing image reconstruction and image editability, HyperStyle allows for accurate and flexible modifications of real images. This makes it a powerful tool for tasks that require fine-grained control over image attributes, such as generating synthetic variations of medical images for training data augmentation. Together, these models provide a robust framework for our unsupervised transformation pipeline, enabling us to generate high-quality synthetic images using inverted codes from images obtained in the real world.

Towards the final step of controlled augmentation generation, existing medical imaging research for developing transformations in the GAN latent space predominantly rely on classification models. While these models have been instrumental in driving advances in image synthesis and manipulation, they come with significant drawbacks, as mentioned in ~\Cref{sec:intro}. To address these concerns, we explore factorizing the latent space of the generator model as an alternative approach to extract semantics in an unsupervised manner. Our proposed pipeline significantly reduces the dependency on scarce and costly labeled data. This unsupervised approach is inherently more scalable, as it can leverage vast amounts of unlabeled data, which is more readily available, thus facilitating the training of models on a broader spectrum of semantic variations.

The overall augmentation pipeline, detailed in \Cref{subsec:unsupervised_transformation_pipeline}, progresses step-by-step from a set of original images to the final outputs, incorporating semantic variations based on semi-automatically extracted features into the original images. First, we apply closed-form factorization to identify meaningful and orthogonal latent semantic directions within the latent space. Next, we utilize the GAN inversion function to map real images into the latent space accurately. Finally, using the semantic directions extracted through factorization, we produce new variants of the original images based on the identified semantics. This approach allows for the exploration of a broad range of semantic variations without the need for labeled data, ensuring that the synthetic outputs closely resemble the original inputs.

\subsection{Unsupervised Transformation Pipeline}
\label{subsec:unsupervised_transformation_pipeline} 
\begin{figure}[tb]
  \centering
  \includegraphics[width=0.85\linewidth]{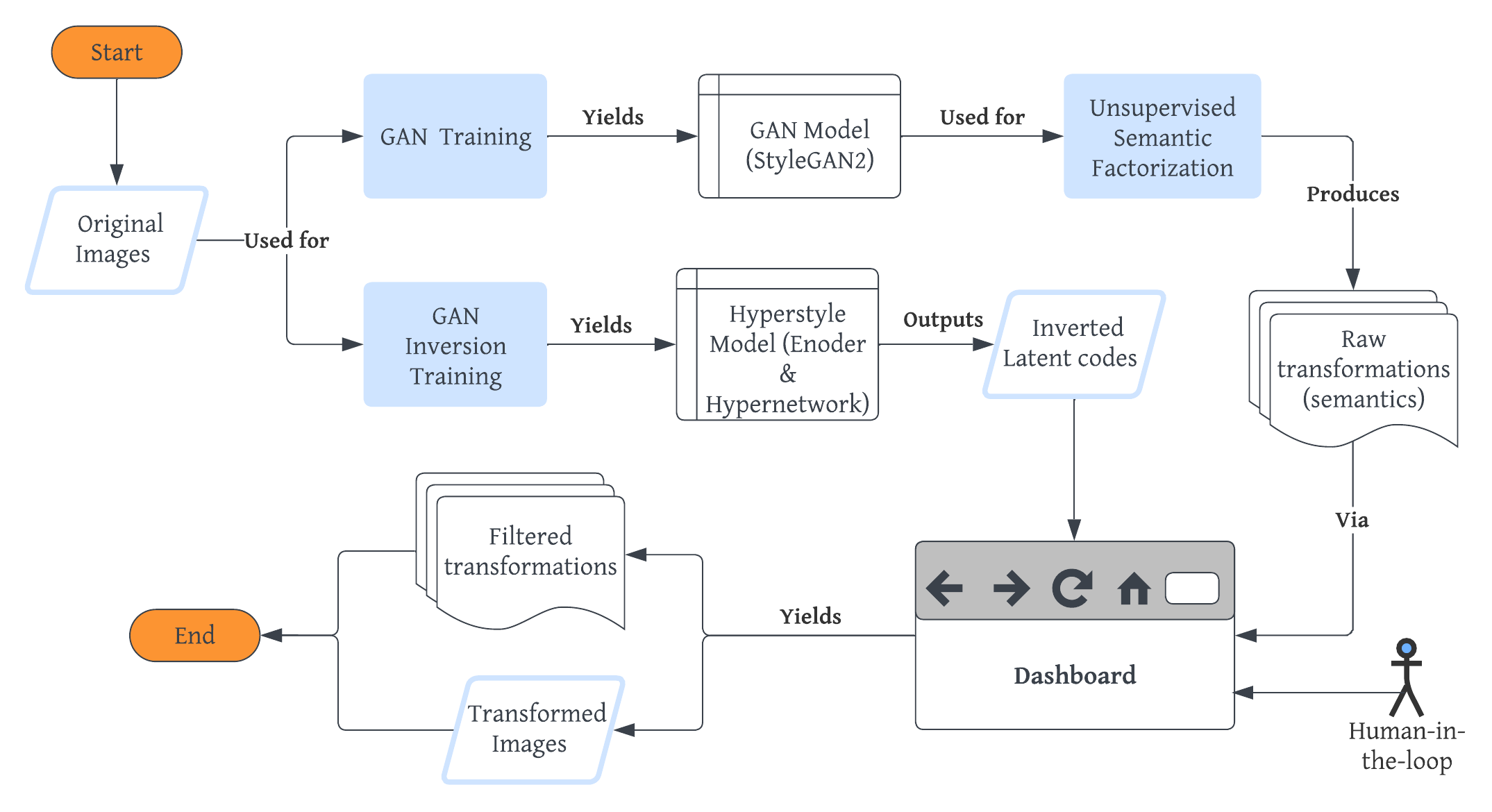}
  \caption{Transformation development pipeline}
  \label{fig:Unsup_Trans_Dev_Pip_Flowchart}
\end{figure}

The proposed transformation development pipeline (see \cref{fig:Unsup_Trans_Dev_Pip_Flowchart}) consists of four stages that perform sequential tasks to achieve the goal of unsupervised semantic extraction. The end product is an unsupervised technique for transformation development, enabling the semi-automatic extraction of extensive semantic transformations reflected within a dataset and corresponding domain.

\begin{enumerate}
\item \textbf{GAN training:} Training a model based on the StyleGAN2 architecture.
\item \textbf{Factorization} to extract eigenvectors of maximal variance (Transformations): This crucial step identifies meaningful, orthogonal latent semantic directions within the latent space with closed-form factorization.
\item \textbf{GAN inversion:} We train a HyperStyle-based GAN inversion model that comprises encoder and optimizer units, with the goal of obtaining latent features corresponding to real images.
\item \textbf{Identify relevant transformations:} The orthogonal latent semantic directions from the previous step correspond to a mix of human-understandable and domain-related concepts. Being able to translate the directions to these concepts contributes to the interpretability of the generated images. Besides, not all the directions produce relevant and unique transformations (i.e., multiple directions may produce very similar transformations). A validation step is incorporated to ensure that only relevant transformations are considered.
\end{enumerate}

In the following sub-sections, we will elaborate on our approach in the context of our case study within the dermatoscopy domain.

\subsubsection{GAN Training}
\label{subsubsec:GAN_training}
We trained the GAN with 10,758 images predominantly from the HAM10000 dataset \cite{tschandl2018ham10000} used in the ISIC 2018 challenge \cite{codella2019skin}. The 10k images from HAM10000 stem from various populations and modalities, with each image annotated with specific diagnoses such as melanoma (MEL), melanocytic nevus (NV), basal cell carcinoma (BCC), actinic keratosis/Bowen’s disease (AKIEC), benign keratosis (BKL), dermatofibroma (DF), and vascular lesion (VASC).

To increase the variability of transformations, we incorporated additional datasets into the HAM10000 dataset. We selected 368 images from the Fitzpatrick dataset \cite{groh2021evaluating}, filtering out images outside the lesion domain; and utilized 390 dermatoscopic images from the Seven-Point Checklist Dermatology dataset \cite{kawahara2019seven}, ensuring that only non-augmented images were included. Additionally, we considered images from the Stanford University dataset \cite{daneshjou2022disparities}, but found that the images contained demarcations and augmentations, leading us to exclude them from our dataset. Demarcated images present visible boundaries and markers that can introduce biases into the training process of the StyleGAN. These markers can disrupt the network's ability to learn the underlying patterns and features of the data, leading to sub-optimal generation of synthetic images. Furthermore, augmentations may alter the natural appearance of the images, causing the model to learn and replicate these alterations rather than the true characteristics of the original images. Therefore, to ensure the integrity and quality of our training data, we opted to exclude this dataset.

We used the StyleGAN2 \cite{karras2020analyzing} architecture for GAN training. We formatted the dataset in LMDB (Lightning Memory-Mapped Database) to take advantage of its speed and low memory usage, which makes it suitable for large-scale data processing. The images were standardized to a fixed resolution of 512 x 512 pixels prior to training to enable an optimal generative quality of the augmented images intended for training lesion classification models at the same resolution.

Through the training process, we fine-tuned the StyleGAN2 model to generate high-quality synthetic skin lesion images. We used the StyleGAN2 architecture, which features redesigned generator normalization, progressive growing, and style-based synthesis blocks to enhance image quality. We kept most of the design details unchanged from the original implementation, including the dimensionality of Z and W spaces (512) and mapping network architecture (8 fully connected layers, 100× lower learning rate). Using Adam optimizer \cite{adamOptimizer} with a learning rate of 0.001 and a batch size of 64, the training spanned 450k iterations, with data augmentation techniques such as random cropping and horizontal flipping to enhance the robustness of the model. We performed the training on a stack of 4 NVIDIA RTX 8000 GPUs in a distributed setup.

\begin{figure}[tb]
  \centering
  \includegraphics[width=.8\linewidth]{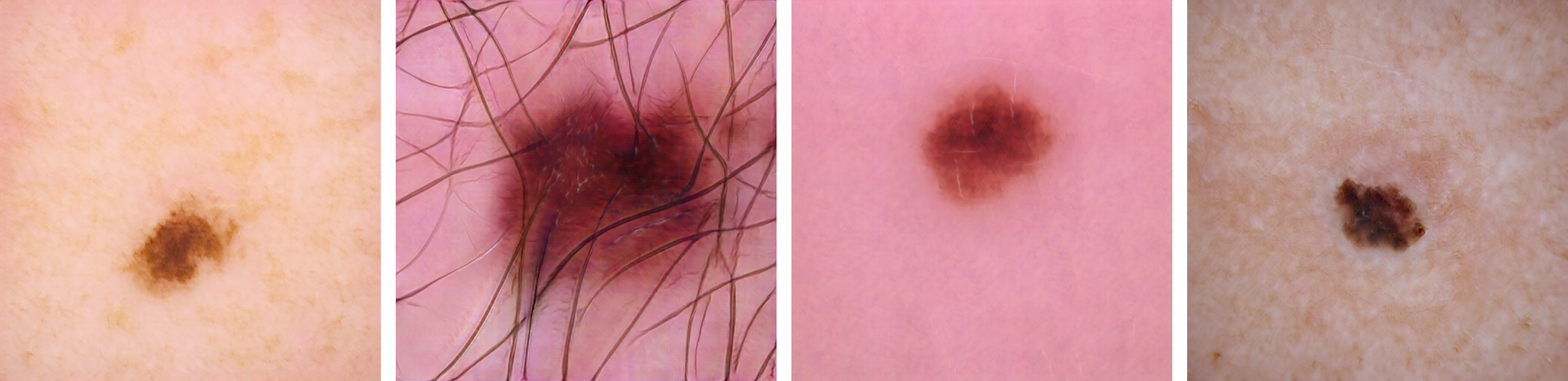}
  \caption{Samples of synthetically generated skin lesions.}
  \label{fig:Generated_Images}
\end{figure}

The model's performance was evaluated using the Fréchet Inception Distance (FID) \cite{heusel2017gans}, which yielded a score of approximately 3.7, indicating a high level of conformance in distributional similarity between the generated and real images. \cref{fig:Generated_Images} showcases samples of synthetically generated dermatoscopic skin images, demonstrating the photorealism achieved by our trained model. 

\subsubsection{Latent-Space Factorization} 
As part of this sub-pipeline, to extract semantic directions or transformations, we employed closed-form factorization\cite{karras2020analyzing} within the latent space (z, w) of the generator to identify meaningful semantic directions. In other words, we analyzed the generator's internal structure to uncover directions in the latent space that correspond to groupings of maximal semantic variance in the generated images.

The process starts with the extraction of specific weights from various layers of the generator. These weights capture the learned features of the model and are critical for generating high-quality images. We then constructed a weight matrix \(W\) that consolidates the weight information from the selected layers. This matrix encapsulates the combined influence of these layers on the generated images.

Next, we applied Singular Value Decomposition (SVD) to the weight matrix \(W\). SVD decomposes \(W\) into three components: \(U\) (an orthogonal matrix), \(E\) (a diagonal matrix containing singular values), and \(V^T\) (the transpose of an orthogonal matrix). The columns of \(V\) (the eigenvectors) represent distinct directions in the latent space along which the data varies the most. These eigenvectors correspond to semantic transformations that can be applied to the latent vectors.

By projecting latent vectors along these eigenvectors, we can create a variety of image transformations the magnitude of which can be varied through the eigenvalue over that direction. Through this process, we were able to uncover transformations for size, texture, geometric properties and background properties of the skin lesion, amongst others. This method allows for extracting subtle variations and intricate features, enhancing the richness and diversity of the synthetic images. This is a pivotal step in our unsupervised transformation development pipeline and empowers testers and domain experts alike to navigate the landscape of transformation variations without relying on extensive manual classification models as discussed earlier.

\subsubsection{GAN Inversion}
As part of the GAN inversion sub-pipeline implementation, we trained a HyperStyle~\cite{alaluf2022hyperstyle} based inversion model comprising dedicated encoder and optimizer units towards the task of latent code (w-space) computation of any image in the real world. For our implementation, we used the e4e (Encoder for Editing) \cite{tov2021designingencoderstyleganimage} encoder, which is a specific type of encoder used to map real images into the latent space of the StyleGAN2 generator.

The e4e encoder was trained on the same dataset used for GAN training in the first phase of the pipeline. The training process involved minimizing the L2 loss (Mean Squared Error), a common metric for measuring the difference between the predicted output of the encoder and the actual target values. We achieved an L2 loss of 0.009, indicating the encoder's success towards distilling and capturing meaningful representations from the dataset.

Following the encoder training, we trained the HyperStyle component using the same dataset. The training process for the HyperStyle module resulted in an L2 loss of 0.002, highlighting its efficacy in refining the latent space for accurate image reconstruction. This low L2 loss value underscores the model's proficiency in transforming latent features into realistic skin lesion images.

\begin{figure}[tb]
  \centering
  \includegraphics[origin=c,width=0.7\linewidth]{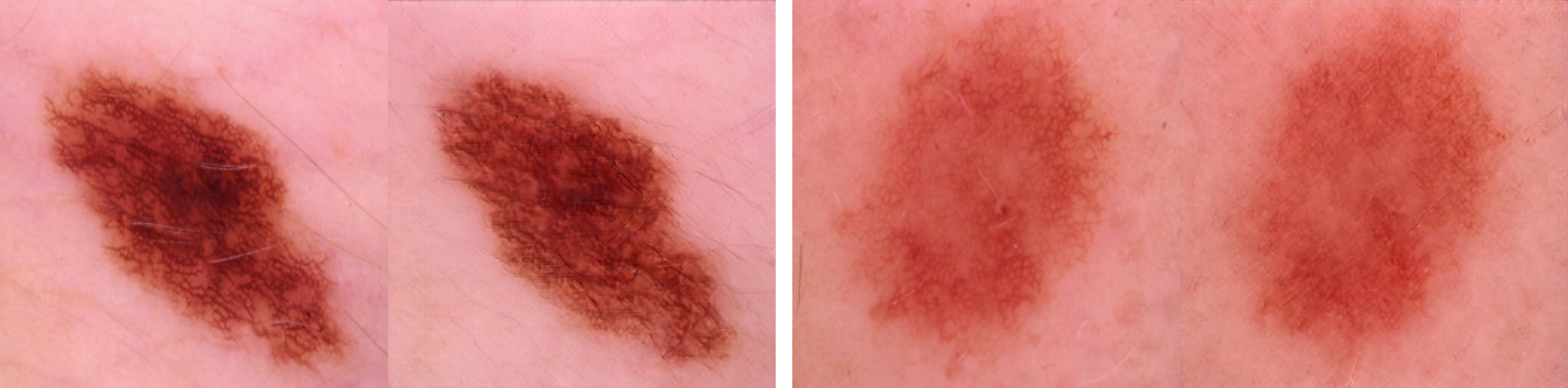}  
  \caption{Two pairs of original (left in each case) and HyperStyle inverted images (right in each case).}
  \label{fig:Inverted_Images}
\end{figure}

\cref{fig:Inverted_Images} showcases the inversion results of several original images. The faithful reconstruction achieved through the synergy between the encoder and HyperStyle components demonstrates the success of the inversion process in capturing intricate semantic properties and detail from the original images.

\subsubsection{Identify Relevant Transformations} 
As part of this phase, we utilized a human-in-the-loop approach to systematically review and validate the semantic directions identified during the closed-form factorization phase. To facilitate this process, we developed a user-friendly dashboard by adapting and adding features to the SeFa (Semantic Factorization) dashboard \cite{sefa}. The dashboard allows the interactive exploration and validation of the semantic transformations. After our modifications, the dashboard supports functionalities such as uploading or browsing images from the dataset, selecting a semantic direction, adjusting the magnitude of the transformation, and visually reviewing the outcome of the applied transformation. This interface is crucial for the interpretation and validation of the semantic meanings and of the relevance of the latent directions identified during factorization (corresponding to this direction and magnitude).

To transition from factorization to identifying transformations, we implemented a method to apply the extracted eigenvectors of maximal variance to the latent vectors of real images. This process involves the following steps:

\begin{enumerate}
    \item We mapped dermatoscopic images into the latent space with the previously trained HyperStyle GAN inversion model.
    \item The identified semantic directions (eigenvectors) are then applied to the latent vectors of these images. By adjusting the magnitude of the directions, we can modulate specific attributes of the images, such as size, pigmentation, and texture of skin lesions.
    \item We used the dashboard to systematically review the transformations, ensuring they are meaningful and relevant to the domain. Multiple directions might produce similar transformations, so this step ensures that only unique and significant transformations are considered.
\end{enumerate}

\begin{figure}[tb]
\centering
    \begin{subfigure}[b]{0.3\textwidth}
        \centering
        \includegraphics[width=\textwidth]{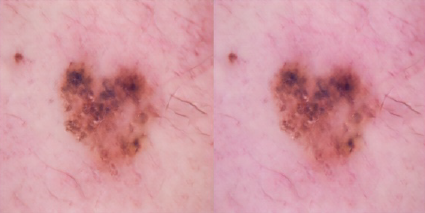}
        \caption{Skin pigmentation change is applied to the original image (left) which resulted in the synthetic image (right).}
    \end{subfigure}
    \hfill
    \begin{subfigure}[b]{0.3\textwidth}
        \centering
        \includegraphics[width=\textwidth]{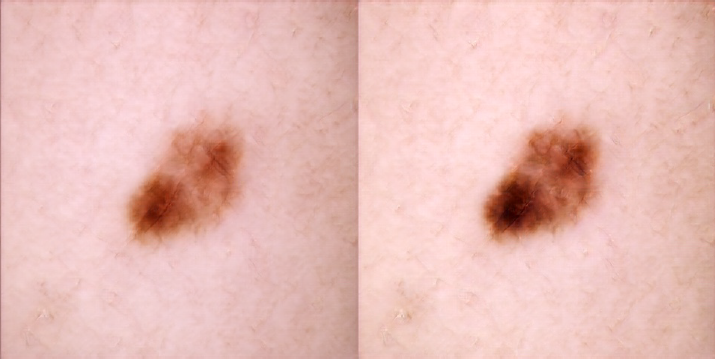}
        \caption{The lesion pigmentation change is applied to the original image (left) which resulted in the synthetic image (right).}
    \end{subfigure}
    \hfill
    \begin{subfigure}[b]{0.3\textwidth}
        \centering
        \includegraphics[width=\textwidth]{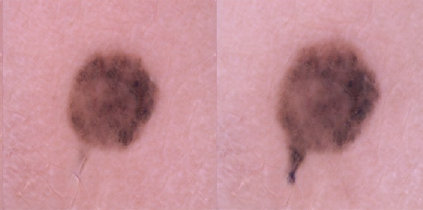}
        \caption{Skin lesion size change is applied to the original image (left) which resulted in the synthetic image (right).}
    \end{subfigure}
    \hfill
    \caption{Examples of original images and their generated/synthetic images after the transformations have been applied to the original images.}
    \label{fig:examples_transformed_images}
\end{figure}

At the end of this process, we identified 13 distinct transformations tailored to the semantic variance permutations of the skin lesion domain. These transformations include changing the skin lesion size, altering the pigmentation of skin lesions, modifying the overall pigmentation of the skin, adjusting the texture and shape of skin lesions, etc. Each transformation represents a modulating force applied to the latent vector of the original image, contributing to a diverse range of augmentation possibilities. This integration of humans-in-the-loop for the semi-automatic semantic extraction is essential for generating diverse transformations that are both relevant and accurate, thereby enabling training robust machine learning models that can generalize well to real-world data. \cref{fig:examples_transformed_images} shows exemplary original images and their corresponding transformations (additional examples of the transformed images are included the Appendix). 

\subsection{Classifier Training Enhancement With Synthetic Data}
\label{subsec:methods_experimental_setup}

\subsubsection{Experimental Setup} 
We compared the predictive performance of a skin lesion classification model trained on data augmented with synthetic images with a baseline. As baseline scenario, we trained on the original HAM10000 \cite{tschandl2018ham10000} dermatoscopy training dataset split and evaluated the performance of the model using the original test dataset split. HAM10000 includes 10,015 high-quality dermatoscopic images in the training set and 1,512 image in the test set. Each image is labeled with one specific diagnosis (see \Cref{subsubsec:GAN_training} above). Class label distribution is highly imbalanced in training and test set, with NV being over-represented with a share of 60~\% respectively 67~\%, while the other six classes share the remaining fraction to varying degrees.

To train the model with additional synthetic images, we first augmented the original training dataset with synthetic data as follows: we generated new images from the original training dataset using five out of thirteen transformations that we had identified. The five transformations correspond to Size and Pigment Variation (SPV), Size Variation (SV), Background Color Variation (BCV), Geometric Variation (GV), and to Positional Shift (PS). The other transformations we excluded were variants of these five transformations (e.g., they correspond to different semantic layers in the generator). 

In total, we obtained five times the amount of the original dataset (total of 50,075 samples). Since we work with a much larger training dataset after the augmentation in comparison to the baseline scenario, we ensured that the observed classification performance change  results not solely from the much larger dataset and thus longer training, but from the higher variety in the training data. For this reason, we employed early stopping (with a relatively high criterion of 25 epochs prior to initiating the early stop; assuring that the model would not profit from longer training) and created multiple ``synthetically augmented'' models by varying the number of synthetic images used for augmenting the training datasets. Specifically, we randomly selected 400, 800, 1200, 1600, and 2000 of the generated synthetic images from each of the five transformations, and augmented the original training dataset with 2000, 4000, 6000, 8000, and 10000 images respectively. Note that the selections of synthetic images for each augmented dataset were done independently; i.e., the 2000 additional images were not a subset to the 4000 additional images. 

To ensure that we are only adding ``good'' synthetic images, we also generated a ``filtered'' augmented training dataset by removing the synthetic images which the unfiltered synthetically augmented models classified incorrectly. We filtered out 136 (6\%), 367 (9\%), 274 (4.5\%), 916 (11.45\%), and 505 (5\%) images from the 2000, 4000, 6000, 8000, 10000 augmented images respectively.

For each augmented training datasets, we trained a classification model. This results in 10 synthetically augmented models: five models were trained using the unfiltered augmented datasets and five models were trained with the filtered augmented datasets. Hereafter, we will refer to the models trained using the unfiltered augmented dataset as SA-2k, SA-4k, SA-6k SA-8k, SA-10k (when the original dataset was augmented with the 2000, 4000, 6000, 8000, and 10000 synthetic images respectively) and the models trained using the filtered augmented dataset as SA-2k-filter, SA-4k-filter, SA-6k-filter, SA-8k-filter and SA-10k-filter.

\subsubsection{Model, Task, and Training} 
We employed a DenseNet121 (8M parameters) and a DenseNet169 (14M parameters), initialized with weights pretrained on ImageNet \cite{5206848}, for multi-class classification. The two architectures enabled us to compare the impact of augmenting the training dataset with synthetic images across varying architecture complexity. Because the label distributions are highly imbalanced, we used weighted oversampling to balance class distributions within training batches. Additional basic transformations (horizontal/vertical flip, cutout) and a dropout rate of 0.1 were employed. We used an Adam optimizer with a learning rate of 1e-5 and weight decay of 1e-4 and trained for 100 epochs, while initiating early stopping when the performance on the validation split did not improve for 25 epochs.

\section{Results and Discussion}

\subsection{Transformations Developed}
To compare our transformed images, we used the \gls{lpips} metric for evaluating the transformed images as it provides a more nuanced 1:1 image-level comparison than FID (which is a measure for comparison of overall image distributions). The \gls{lpips} score we employed is conditioned on the last three layers of the AlexNet architecture \cite{Krizhevsky2012ImageNetCW} trained on the ImageNet dataset and serves to quantify perceptual similarity by comparing deep feature representations extracted across the layers, empirically proven to align with human perceptual judgments. 

We calculated the \gls{lpips} metrics for the five transformations used in augmenting the training datasets (see \cref{tab:transformation_metrics}.) \gls{lpips} score ranges between 0 to 1 where a lower \gls{lpips} score denotes higher perceptual similarity.

We observe scores close to or lower than 0.1 for all our selected transformations. In general, ``wayward or low-fidelity'' transformations exhibited scores > 0.2, which was the threshold used in selecting transformations for our task. Although the scores for our selected transformations show minimal perceptual change, we acknowledge the importance of domain expert validation to enhance confidence in the fidelity of our transformed images (note that downstream classifiers were always tested on non-modified images). Additionally, in future work, we intend to condition the metric on an AlexNet architecture trained specifically on an unbiased skin lesion dataset to ensure higher resonance in comparison over the feature space. 

\begin{table}[tb]
    \caption{\gls{lpips} metrics for the five transformations employed in training the classification model.}
    \label{tab:transformation_metrics}
    \centering
    \setlength{\tabcolsep}{6pt}
    \begin{tabular}{c | ccccc}
        \toprule
        {Transformation} & {SPV} & {SV} & {BCV} & {GV} & {PS} \\
        \midrule
        {\gls{lpips}} & 0.101 & 0.098 & 0.098 & 0.098 & 0.099 \\
        \bottomrule
    \end{tabular}

\end{table}

\subsection{Evaluation}
We evaluated the model performance on the 1,512 images (512 x 512 pixels) of the original HAM10000 test split, which were neither transformed nor seen during training, using balanced multi-class accuracy.
First, we compared our results with the existing benchmark\cite{ISIC2018_leaderboard} (`Task 3: Lesion Diagnosis') in the ISIC2018 challenge. Our best performing model was based on the DenseNet169 architecture, synthetically augmented with 6000 additional synthetic images (60 \% of the original training dataset), achieved a balanced accuracy of 0.856 (see \Cref{tab:synth_aug_vs_baseline}). Comparing with other models evaluated in the challenge \cite{ISIC2018_leaderboard}, we ranked 3rd on the evaluation metrics, with only the two ensemble based methods achieving a higher average balanced accuracy of 0.885 (`Top 10 Models Averaged') \cite{nozdryn2018ensembling} and 0.856 (`Large Ensemble with heavy multi-cropping and loss weighting') \cite{gessert2018skinlesiondiagnosisusing} respectively. 
Our model even surpasses the larger DenseNet201 on rank 4 in the challenge with 0.815 (`densenet' submitted by Li and Li \cite{ISIC2018_leaderboard}).

Then, we compared the performance of the synthetically augmented classification models with the baseline model. 
In \cref{tab:synth_aug_vs_baseline}, we can observe that all synthetically augmented models outperform the baseline model by 1.9\% to 3.5\%. However, the performance does not always increase with the number of synthetic images added to the original dataset and seems to plateau after adding 6,000 images to the original training dataset. We also observe that the filtered method helps to increase the performance gain only to a certain point. This suggests that more research is needed to understand the nature of the synthetic images that increase and/or decrease the models' performance.

\begin{table}[tb]
    \caption{Classification performance improvement of the synthetically augmented models in comparison to the baseline model (trained with real training data only), measured with weighted multi-class accuracy.
    }
    \label{tab:synth_aug_vs_baseline}
    \setlength{\tabcolsep}{6pt}
    \centering
    \begin{tabular}{c | c| c| c | c | c | c | c }
    \toprule
       Architecture & Filter & Baseline & 2k & 4k & 6k & 8k & 10k  \\
        \midrule
        DenseNet121 & No  & 81.9\% & +1.9\% & +2.2\% & +2.6\% & +1.2\% & +1.8\% \\
        DenseNet169 & No  & 82.1\% & +1.9\% & +2.2\% & +3.5\% & +3.3\% & +3.1\% \\
        DenseNet169 & Yes & 82.1\% & +2.3\% & +3.3\% & +3.5\% & +3.1\% & +2.8\% \\ \bottomrule
    \end{tabular}
\end{table}

\begin{figure}[tb]
\centering
    \begin{subfigure}[b]{0.46\textwidth}
        \centering
        \includegraphics[width=\textwidth]{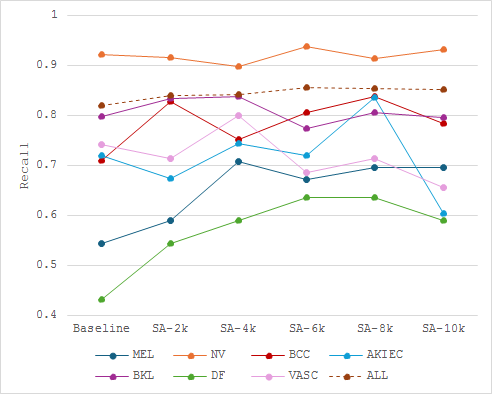}
        \caption{Models trained on \textbf{unfiltered} augmented datasets of varying size compared to baseline.}
    \end{subfigure}
    \hfill
    \begin{subfigure}[b]{0.46\textwidth}
        \centering
        \includegraphics[width=\textwidth]{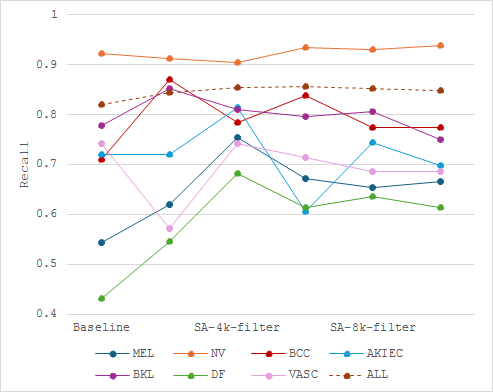}
        \caption{Models trained on \textbf{filtered} augmented datasets of varying size compared to baseline.}
    \end{subfigure}
    \caption{Predictive performance (recall) of the synthetically augmented models compared to the baseline model.}
    \label{fig:model_performances}
\end{figure}

\begin{figure}[tb]
    \centering
    \begin{subfigure}[b]{0.32\textwidth}
        \centering
        \includegraphics[width=\textwidth]{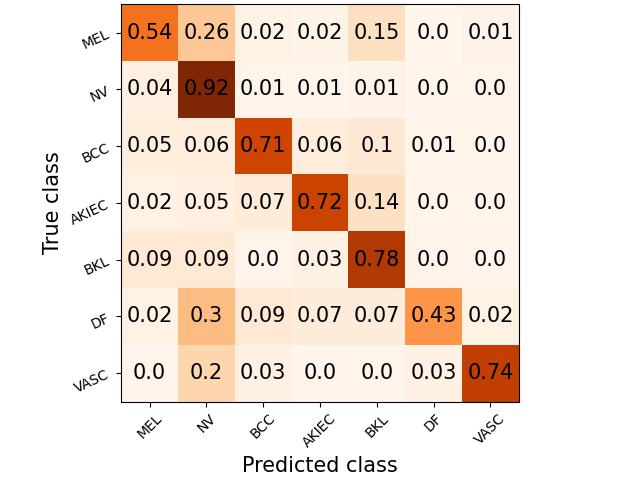}
        \caption{Baseline}
        \label{fig:fig1}
    \end{subfigure}
    \hfill
    \begin{subfigure}[b]{0.32\textwidth}
        \centering
        \includegraphics[width=\textwidth]{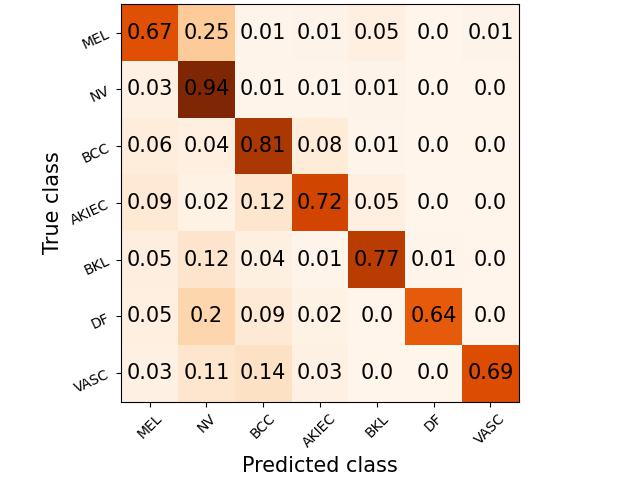}
        \caption{SA-6K model}
        \label{fig:fig2}
    \end{subfigure}
    \hfill
    \begin{subfigure}[b]{0.32\textwidth}
        \centering
        \includegraphics[width=\textwidth]{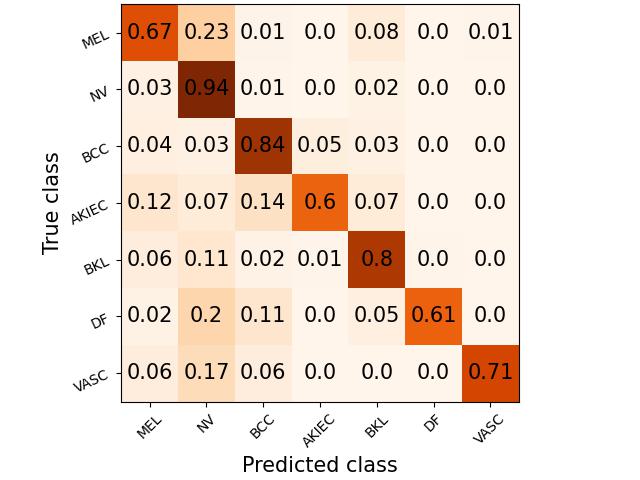}
        \caption{SA-6k-filter model}
        \label{fig:fig3}
    \end{subfigure}
    \caption{Confusion matrices, baseline vs. synthetically augmented un/filtered models.}
    \label{fig:confusion_matrix}
\end{figure}

From here on we will report only on the DenseNet169, as it consistently outperforms the smaller DenseNet121. \cref{fig:model_performances} shows the recall results for each diagnostic class for the un/filtered synthetically augmented models in comparison to the baseline model. 
Synthetically augmented (un/filtered) models show a better recall for the classes MEL, BCC, and DF, when considering all sizes of augmented data. Note that all three are underrepresented classes, which shows our synthetic augmentations are fit to counteract class imbalances. The synthetically augmented models have better recall performance for all classes, except for BKL and VASC, when only considering the SA-6k and SA-6k-filter. \cref{fig:confusion_matrix} shows the confusion matrices for the baseline, un/filtered synthetically augmented models (using 6000 augmented images). \cref{tab:AUCROC} shows the area under the curve of the receiver operating characteristic (AUC ROC) for each class for the same models. The table demonstrates that both optimized models (SA-6k and SA-6k-filter) outperform the baseline model in almost all classes. Furthermore, the average AUC ROC for SA-6k is 0.945, higher than the baseline's 0.924, indicating a general performance boost. Meanwhile, the average AUC ROC for SA-6k-filter is the highest at 0.947, suggesting it is the most effective model overall. This indicates that the additional enhancements and filtering techniques applied in SA-6k-filter lead to the most reliable and accurate model for distinguishing between different types of skin lesions. 

\begin{table}[tb]
    \caption{AUC ROC per class for baseline, and un/filtered models (best in bold font).}
    \label{tab:AUCROC}
    \centering
    \setlength{\tabcolsep}{5pt}
    \begin{tabular}{c|ccccccc|c}
        \toprule
        {} & {MEL}  & {NV} & {BCC} & {AKIEC} & {BKL} & {DF} & {VASC} & {Average} \\
        \midrule
        Baseline & 0.832 & 0.938 & 0.961 & 0.951 & \textbf{0.945} & 0.891 & \textbf{0.950} & 0.924\\
        SA-6k & \textbf{0.908} & 0.948 & 0.972 & \textbf{0.973} & 0.937 & 0.930 & 0.944 & 0.945\\
        SA-6k-filter & 0.893 & \textbf{0.952} & \textbf{0.976} & 0.964 & 0.938 & \textbf{0.956} & \textbf{0.950} & \textbf{0.947}\\
        \bottomrule
    \end{tabular}
\end{table}

\subsection{Model Analysis with Explainable AI}
\begin{figure}[h!]
    \centering
    \begin{subfigure}[b]{0.4\textwidth}
        \centering
        \includegraphics[width=\textwidth]{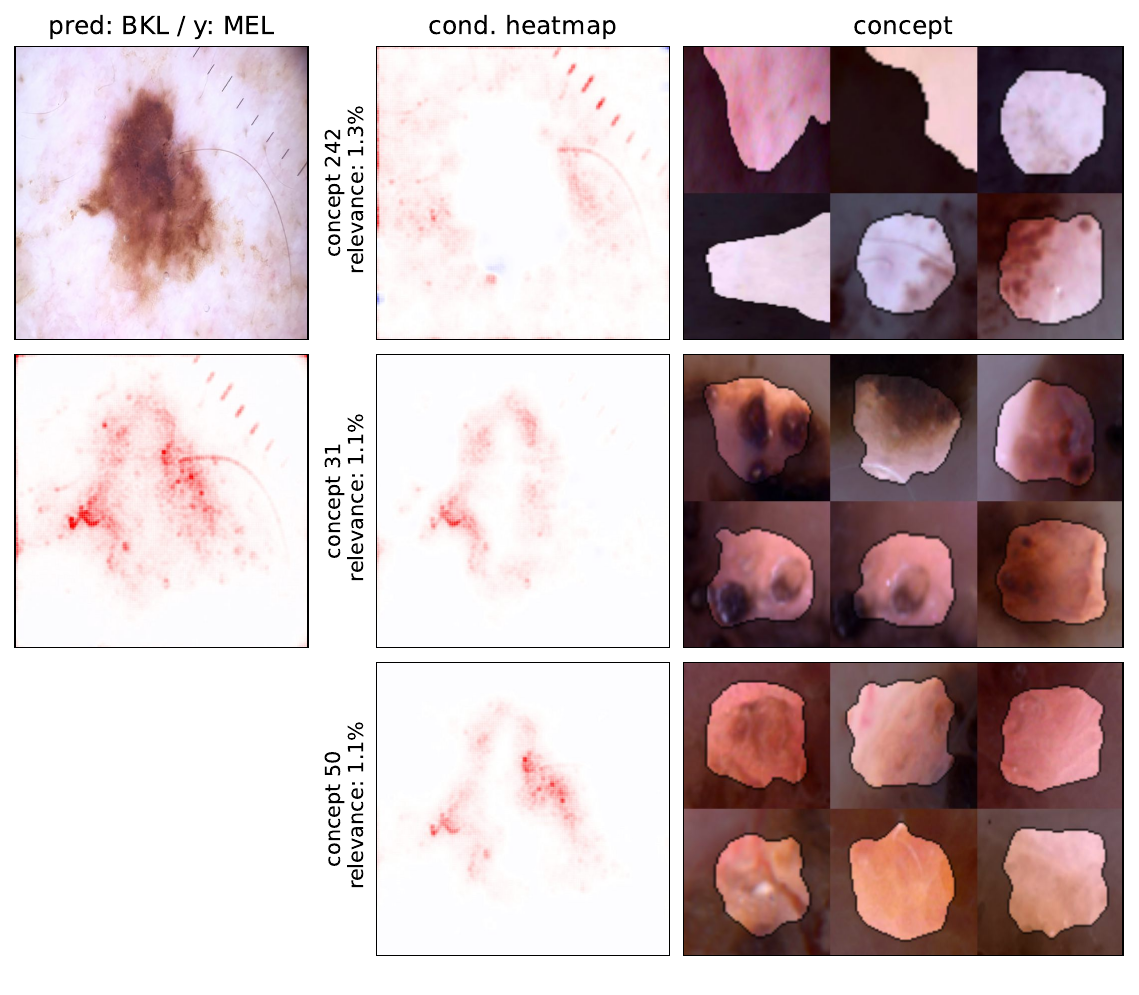}
        \caption{Baseline}
        \label{fig:fig4}
    \end{subfigure}
    \begin{subfigure}[b]{0.4\textwidth}
        \centering
        \includegraphics[width=\textwidth]{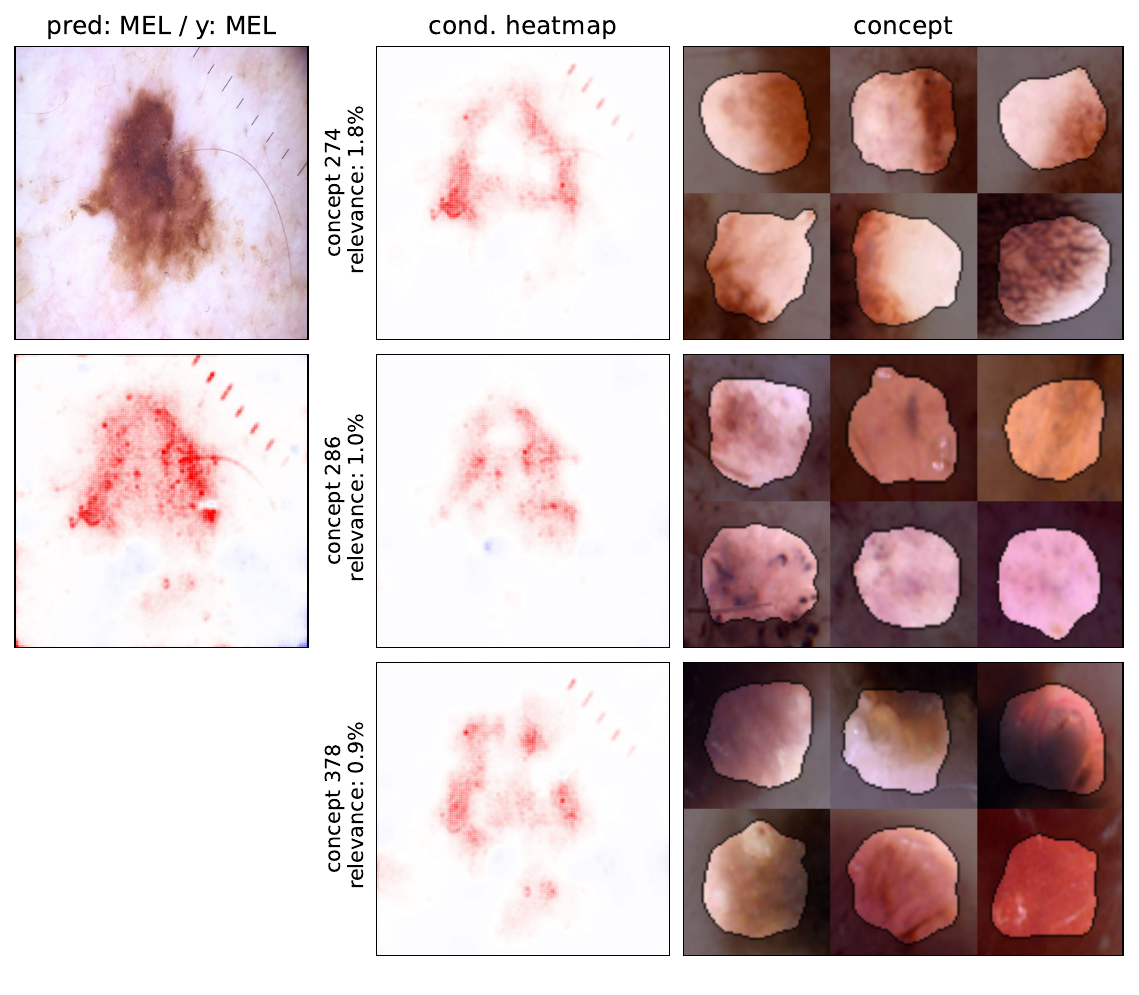}
        \caption{Synthetic (Filtered)}
        \label{fig:fig5}
    \end{subfigure}

    \caption{CRP analysis for a test sample misclassified as BKL by the baseline model (\emph{left}), but correctly classified as MEL by the augmented model (\emph{right}): We show concept-conditional heatmaps for the most relevant concepts for the predictions and concept visualizations with a set of reference images. For interpretability, we zoom into the most relevant regions of reference samples and mask out irrelevant areas.}
    \label{fig:crp_analysis}
    %\vspace{-15pt}

\end{figure}

The confusion matrices in Fig.~\ref{fig:confusion_matrix} reveal a significant improvement of the model's ability to correctly classify samples from class MEL. 
Whereas the baseline model only classified 54\% of true Melanoma samples correctly, the model trained on (filtered) synthetic samples correctly labeled 67\% of these samples.
As the baseline model misclassified many true MEL test samples as BKL, which is particularly dangerous as this is a benign class, we further analyze the prediction behavior for this set of test samples. 
Specifically, we apply \gls{crp}~\cite{achtibat2023attribution} to compute concept-based explanations for individual predictions. 
\gls{crp} disentangles local explanations into concept-specific explanations.
The concepts are defined by individual neurons in a chosen layer (e.g., last Conv layer) and their relevance scores can be computed with backpropagation-based explainers, for instance \gls{lrp}~\cite{bach2015pixel}.
The concepts can be visualized in a human-understandable manner by a set of representative samples from a reference dataset, e.g., the training data. 
Fig.~\ref{fig:crp_analysis} shows \gls{crp} explanations for a test sample misclassified as BKL by the baseline model (\emph{left}) but classified correctly by the model augmented with synthetic data (\emph{right}). 
While the baseline model is distracted by surroundings (e.g., concept 242), the augmented model uses features easier to interpret and more related to the task, such as the border of the mole (concept 274). We include Figure \ref{fig:app:crp_groundtruth} in the appendices, which shows explanations of the MEL class for the baseline model. These explanations reveal that the baseline model is not capable of detecting any interesting features which indicate membership to the MEL class, as opposed to the augmented model. Furthermore, to understand the global prediction (sub-)strategies employed by the model, we compute \gls{pcx}~\cite{dreyer2024understanding} for class MEL. These explanations can be found in Appendix \ref{appendix:pcx_section}.

\section{Conclusion and Future Work}

Our research has established a robust foundation for the efficient and effective utilization of controlled augmentation using generative models to generate synthetic skin lesion images and consequently more accurate AI classification models. Through our experiments, we have demonstrated significant improvements in model performance for skin lesion classification by using augmented datasets generated in a controlled manner with our implementation. In these experiments, the selection of augmented datasets for training was done by randomly sampling from synthetically generated images. To enhance the efficiency of this approach, in future work, we plan to modify our data selection strategy to be based on clustering characteristics within the latent space, thereby selecting images from which the model stands to learn the most. Additionally, we intend to implement a filtration module prior to augmented data selection, based on the development of fidelity and photorealism metrics and thresholds. To achieve this, we will build on the Learned Perceptual Image Patch Similarity (LPIPS) metric and aim to establish thresholds for the photorealism and fidelity of the augmented datasets selected for model training, thereby preventing unintentional data poisoning.%In our exploration of image inversion, we have successfully used HyperStyle as an efficient technique for balancing the trade-off between reconstruction and editability. However, 

We plan to further explore the trade-off between photorealism and editability and investigate other inversion techniques to improve the optimization process. Additionally, future work will focus on enhancing our factorization techniques to produce high-fidelity directions of disentangled semantic variance. While our results currently lead the ISIC leaderboard for non-ensemble-based models, we believe that by shifting to an ensemble-based approach, we can surpass the performance of the leading ensemble-based models. 
We believe that our research can pave the way for future advancements in transfer learning and domain adaptation within dermatological diagnoses. We will further explore generalizability to other diagnostic tasks and datasets, as well as higher-dimensional image analysis such as hyperspectral tissue differentiation.%\cite{wisotzky2024}.

%References should start immediately after the main text, but can continue past p.\ 14 if needed.
%\clearpage  % TODO REVIEW/FINAL: This \clearpage needs to be removed from both review and camera-ready versions.

\bibliographystyle{splncs04}
\bibliography{references}

\begin{thebibliography}{10}
\providecommand{\url}[1]{\texttt{#1}}
\providecommand{\urlprefix}{URL }
\providecommand{\doi}[1]{https://doi.org/#1}

\bibitem{ISIC2018_leaderboard}
{ISIC} challenge. \url{https://challenge.isic-archive.com/leaderboards/2018/}

\bibitem{abhishek2019mask2lesion}
Abhishek, K., Hamarneh, G.: Mask2{L}esion: Mask-constrained adversarial skin lesion image synthesis (2019), \url{https://arxiv.org/abs/1906.05845}

\bibitem{achtibat2023attribution}
Achtibat, R., Dreyer, M., Eisenbraun, I., Bosse, S., Wiegand, T., Samek, W., Lapuschkin, S.: From attribution maps to human-understandable explanations through concept relevance propagation. Nat. Mach. Intell.  \textbf{5}(9),  1006--19 (2023)

\bibitem{akrout2023diffusion}
Akrout, M., Gyepesi, B., Holl{\'o}, P., Po{\'o}r, A., Kincs{\H{o}}, B., Solis, S., Cirone, K., Kawahara, J., Slade, D., Abid, L., Kov{\'a}cs, M., Fazekas, I.: Diffusion-based data augmentation for skin disease classification: Impact across original medical datasets to fully synthetic images. In: Mukhopadhyay, A., Oksuz, I., Engelhardt, S., Zhu, D., Yuan, Y. (eds.) Deep Generative Models. pp. 99--109. Springer, Cham (2024). \doi{10.1007/978-3-031-53767-7_10}

\bibitem{alaluf2022hyperstyle}
Alaluf, Y., Tov, O., Mokady, R., Gal, R., Bermano, A.: {HyperStyle}: {StyleGAN} inversion with hypernetworks for real image editing. In: Proc. IEEE/CVF CVPR. pp. 18511--18521. IEEE, NYC, USA (June 2022)

\bibitem{aversa2023diffinfinite}
Aversa, M., Nobis, G., H\"{a}gele, M., Standvoss, K., Chirica, M., Murray-Smith, R., Alaa, A.M., Ruff, L., Ivanova, D., Samek, W., Klauschen, F., Sanguinetti, B., Oala, L.: {DiffInfinite}: Large mask-image synthesis via parallel random patch diffusion in histopathology. In: Oh, A., Neumann, T., Globerson, A., Saenko, K., Hardt, M., Levine, S. (eds.) Advances in Neural Information Processing Systems. vol.~36, pp. 78126--78141. Curran Assoc., Inc., NYC, USA (2023)

\bibitem{bach2015pixel}
Bach, S., Binder, A., Montavon, G., Klauschen, F., M{\"u}ller, K.R., Samek, W.: On pixel-wise explanations for non-linear classifier decisions by layer-wise relevance propagation. PloS one  \textbf{10}(7),  e0130140 (2015)

\bibitem{baur2018generating}
Baur, C., Albarqouni, S., Navab, N.: Generating Highly Realistic Images of Skin Lesions with GANs, p. 260–267. Springer, London, UK (2018). \doi{10.1007/978-3-030-01201-4_28}

\bibitem{baur2018melanogans}
Baur, C., Albarqouni, S., Navab, N.: {Melano{GAN}s}: High resolution skin lesion synthesis with {GANs} (2018), \url{https://arxiv.org/abs/1804.04338}

\bibitem{brinker2019deep}
Brinker, T.J., Hekler, A., Enk, A.H., Klode, J., Hauschild, A., Berking, C., Schilling, B., Haferkamp, S., Schadendorf, D., Holland-Letz, T., et~al.: Deep learning outperformed 136 of 157 dermatologists in a head-to-head dermoscopic melanoma image classification task. Eur. J. Cancer  \textbf{113},  47--54 (2019)

\bibitem{chen2022generative}
Chen, Y., Yang, X.H., Wei, Z., Heidari, A.A., Zheng, N., Li, Z., Chen, H., Hu, H., Zhou, Q., Guan, Q.: Generative adversarial networks in medical image augmentation: A review. Comput. Biol. Med.  \textbf{144},  105382 (2022). \doi{10.1016/j.compbiomed.2022.105382}

\bibitem{codella2019skin}
Codella, N., Rotemberg, V., Tschandl, P., Celebi, M.E., Dusza, S., Gutman, D., Helba, B., Kalloo, A., Liopyris, K., Marchetti, M., Kittler, H., Halpern, A.: Skin lesion analysis toward melanoma detection 2018: A challenge hosted by the international skin imaging collaboration ({ISIC}) (2019)

\bibitem{daneshjou2022disparities}
Daneshjou, R., Vodrahalli, K., Novoa, R.A., Jenkins, M., Liang, W., Rotemberg, V., Ko, J., Swetter, S.M., Bailey, E.E., Gevaert, O., Mukherjee, P., Phung, M., Yekrang, K., Fong, B., Sahasrabudhe, R., Allerup, J.A.C., Okata-Karigane, U., Zou, J., Chiou, A.S.: Disparities in dermatology {AI} performance on a diverse, curated clinical image set. Science Advances  \textbf{8}(32),  eabq6147 (2022). \doi{10.1126/sciadv.abq6147}

\bibitem{5206848}
Deng, J., Dong, W., Socher, R., Li, L.J., Li, K., Fei-Fei, L.: Imagenet: A large-scale hierarchical image database. In: 2009 IEEE Conference on Computer Vision and Pattern Recognition. pp. 248--255 (2009). \doi{10.1109/CVPR.2009.5206848}

\bibitem{dolezal2023deep}
Dolezal, J.M., Wolk, R., Hieromnimon, H.M., Howard, F.M., Srisuwananukorn, A., Karpeyev, D., Ramesh, S., Kochanny, S., Kwon, J.W., Agni, M., Simon, R.C., Desai, C., Kherallah, R., Nguyen, T.D., Schulte, J.J., Cole, K., Khramtsova, G., Garassino, M.C., Husain, A.N., Li, H., Grossman, R., Cipriani, N.A., Pearson, A.T.: Deep learning generates synthetic cancer histology for explainability and education. npj Precis. Oncol.  \textbf{7}(1), ~49 (May 2023). \doi{10.1038/s41698-023-00399-4}

\bibitem{dreyer2024understanding}
Dreyer, M., Achtibat, R., Samek, W., Lapuschkin, S.: Understanding the (extra-) ordinary: Validating deep model decisions with prototypical concept-based explanations. In: Proceedings of the IEEE/CVF Conference on Computer Vision and Pattern Recognition. pp. 3491--3501 (2024)

\bibitem{gessert2018skinlesiondiagnosisusing}
Gessert, N., Sentker, T., Madesta, F., Schmitz, R., Kniep, H., Baltruschat, I., Werner, R., Schlaefer, A.: Skin lesion diagnosis using ensembles, unscaled multi-crop evaluation and loss weighting (2018), \url{https://arxiv.org/abs/1808.01694}

\bibitem{goodfellow2014generative}
Goodfellow, I., Pouget-Abadie, J., Mirza, M., Xu, B., Warde-Farley, D., Ozair, S., Courville, A., Bengio, Y.: Generative adversarial nets. Advances in neural information processing systems  \textbf{27} (2014)

\bibitem{groh2021evaluating}
Groh, M., Harris, C., Soenksen, L., Lau, F., Han, R., Kim, A., Koochek, A., Badri, O.: Evaluating deep neural networks trained on clinical images in dermatology with the {F}itzpatrick 17k dataset. In: 2021 IEEE/CVF CVPRW. pp. 1820--1828. IEEE, NYC, USA (2021). \doi{10.1109/CVPRW53098.2021.00201}

\bibitem{heusel2017gans}
Heusel, M., Ramsauer, H., Unterthiner, T., Nessler, B., Hochreiter, S.: {GAN}s trained by a two time-scale update rule converge to a local {N}ash equilibrium. Advances in Neural Information Processing Systems (NIPS 2017)  \textbf{30} (2017)

\bibitem{jeong2022systematic}
Jeong, J.J., Tariq, A., Adejumo, T., Trivedi, H., Gichoya, J.W., Banerjee, I.: Systematic review of generative adversarial networks ({GANs}) for medical image classification and segmentation. J. Digit. Imaging  \textbf{35}(2),  137--152 (2022). \doi{https://doi.org/10.1007/s10278-021-00556-w}

\bibitem{karras2020analyzing}
Karras, T., Laine, S., Aittala, M., Hellsten, J., Lehtinen, J., Aila, T.: Analyzing and improving the image quality of {StyleGAN}. In: 2020 IEEE/CVF CVPR. pp. 8107--8116. IEEE, NYC, USA (2020). \doi{10.1109/CVPR42600.2020.00813}

\bibitem{kawahara2019seven}
Kawahara, J., Daneshvar, S., Argenziano, G., Hamarneh, G.: Seven-point checklist and skin lesion classification using multitask multimodal neural nets. IEEE J. Biomed. Health Inform.  \textbf{23}(2),  538--546 (2019). \doi{10.1109/JBHI.2018.2824327}

\bibitem{kebaili2023deep}
Kebaili, A., Lapuyade-Lahorgue, J., Ruan, S.: Deep learning approaches for data augmentation in medical imaging: a review. J. Imaging  \textbf{9}(4), ~81 (2023). \doi{10.3390/jimaging9040081}

\bibitem{adamOptimizer}
Kingma, D.P., Ba, J.: Adam: A method for stochastic optimization (2017), \url{https://arxiv.org/abs/1412.6980}

\bibitem{Krizhevsky2012ImageNetCW}
Krizhevsky, A., Sutskever, I., Hinton, G.E.: Imagenet classification with deep convolutional neural networks. Communications of the ACM  \textbf{60},  84 -- 90 (2012), \url{https://api.semanticscholar.org/CorpusID:195908774}

\bibitem{levine2020synthesis}
Levine, A.B., Peng, J., Farnell, D., Nursey, M., Wang, Y., Naso, J.R., Ren, H., Farahani, H., Chen, C., Chiu, D., Talhouk, A., Sheffield, B., Riazy, M., Ip, P.P., Parra‐Herran, C., Mills, A., Singh, N., Tessier‐Cloutier, B., Salisbury, T., Lee, J., Salcudean, T., Jones, S.J., Huntsman, D.G., Gilks, C.B., Yip, S., Bashashati, A.: Synthesis of diagnostic quality cancer pathology images by generative adversarial networks. J. Pathol.  \textbf{252}(2),  178–188 (Sep 2020). \doi{10.1002/path.5509}

\bibitem{makhlouf2023use}
Makhlouf, A., Maayah, M., Abughanam, N., Catal, C.: The use of generative adversarial networks in medical image augmentation. Neural Comput. Appl.  \textbf{35}(34),  24055--24068 (2023). \doi{10.1007/s00521-023-09100-z}

\bibitem{nozdryn2018ensembling}
Nozdryn-Plotnicki, A., Yap, J., Yolland, W.: Ensembling convolutional neural networks for skin cancer classification. In: International Skin Imaging Collaboration (ISIC) Challenge on Skin Image Analysis for Melanoma Detection, MICCAI (2018)

\bibitem{qasim2020red-gan}
Qasim, A.B., Ezhov, I., Shit, S., Schoppe, O., Paetzold, J.C., Sekuboyina, A., Kofler, F., Lipkova, J., Li, H., Menze, B.: {Red-GAN}: Attacking class imbalance via conditioned generation. {Y}et another medical imaging perspective. In: Arbel, T., Ben~Ayed, I., de~Bruijne, M., Descoteaux, M., Lombaert, H., Pal, C. (eds.) Proc. Third Conference on Medical Imaging with Deep Learning. Proc. Machine Learning Research, vol.~121, pp. 655--668. PMLR (06--08 Jul 2020), \url{https://proceedings.mlr.press/v121/qasim20a.html}

\bibitem{quiros2021pathologyGAN}
Quiros, A.C., Murray-Smith, R., Yuan, K.: Pathology{GAN}: Learning deep representations of cancer tissue. Machine Learning for Biomedical Imaging  \textbf{1},  1--47 (2021). \doi{10.59275/j.melba.2021-gfgg}

\bibitem{sagers2022improving}
Sagers, L.W., Diao, J.A., Groh, M., Rajpurkar, P., Adamson, A., Manrai, A.K.: Improving dermatology classifiers across populations using images generated by large diffusion models. In: NeurIPS 2022 Workshop on Synthetic Data for Empowering ML Research (2022), \url{https://openreview.net/forum?id=Vzdbjtz6Tys}

\bibitem{sefa}
Shen, Y., Zhou, B.: Closed-form factorization of latent semantics in {GAN}s. CoRR  (2020), \url{https://arxiv.org/abs/2007.06600}

\bibitem{solanki2023gans}
Solanki, A., Naved, M.: GANs for Data Augmentation in Healthcare. Springer, Cham (2023). \doi{10.1007/978-3-031-43205-7}

\bibitem{tov2021designingencoderstyleganimage}
Tov, O., Alaluf, Y., Nitzan, Y., Patashnik, O., Cohen-Or, D.: Designing an encoder for {S}tyle{GAN} image manipulation (2021), \url{https://arxiv.org/abs/2102.02766}

\bibitem{tschandl2018ham10000}
Tschandl, P., Rosendahl, C., Kittler, H.: The {HAM}10000 dataset, a large collection of multi-source dermatoscopic images of common pigmented skin lesions. Sci Data  \textbf{5}(1), ~1--9 (2018). \doi{10.1038/sdata.2018.161}

\bibitem{yi2018unsupervised}
Yi, X., Walia, E., Babyn, P.: Unsupervised and semi-supervised learning with categorical generative adversarial networks assisted by {W}asserstein distance for dermoscopy image classification (2018), \url{https://arxiv.org/abs/1804.03700}

\bibitem{yi2019generative}
Yi, X., Walia, E., Babyn, P.: Generative adversarial network in medical imaging: A review. Med. Image Anal.  \textbf{58},  101552 (Dec 2019). \doi{10.1016/j.media.2019.101552}

\end{thebibliography}

\newpage

\renewcommand{\thesection}{\Alph{section}}
\renewcommand\thefigure{\thesection.\arabic{figure}}    
\renewcommand\thetable{\thesection.\arabic{table}}
\renewcommand{\theequation}{\thesection.\arabic{equation}}
\setcounter{figure}{0}   
\setcounter{table}{0}
\setcounter{equation}{0}
\setcounter{section}{0}
\section*{Appendices}

\section{Exemplary transformations}

\begin{figure}[h!]
  \centering
  \includegraphics[origin=c,width=0.85\linewidth]{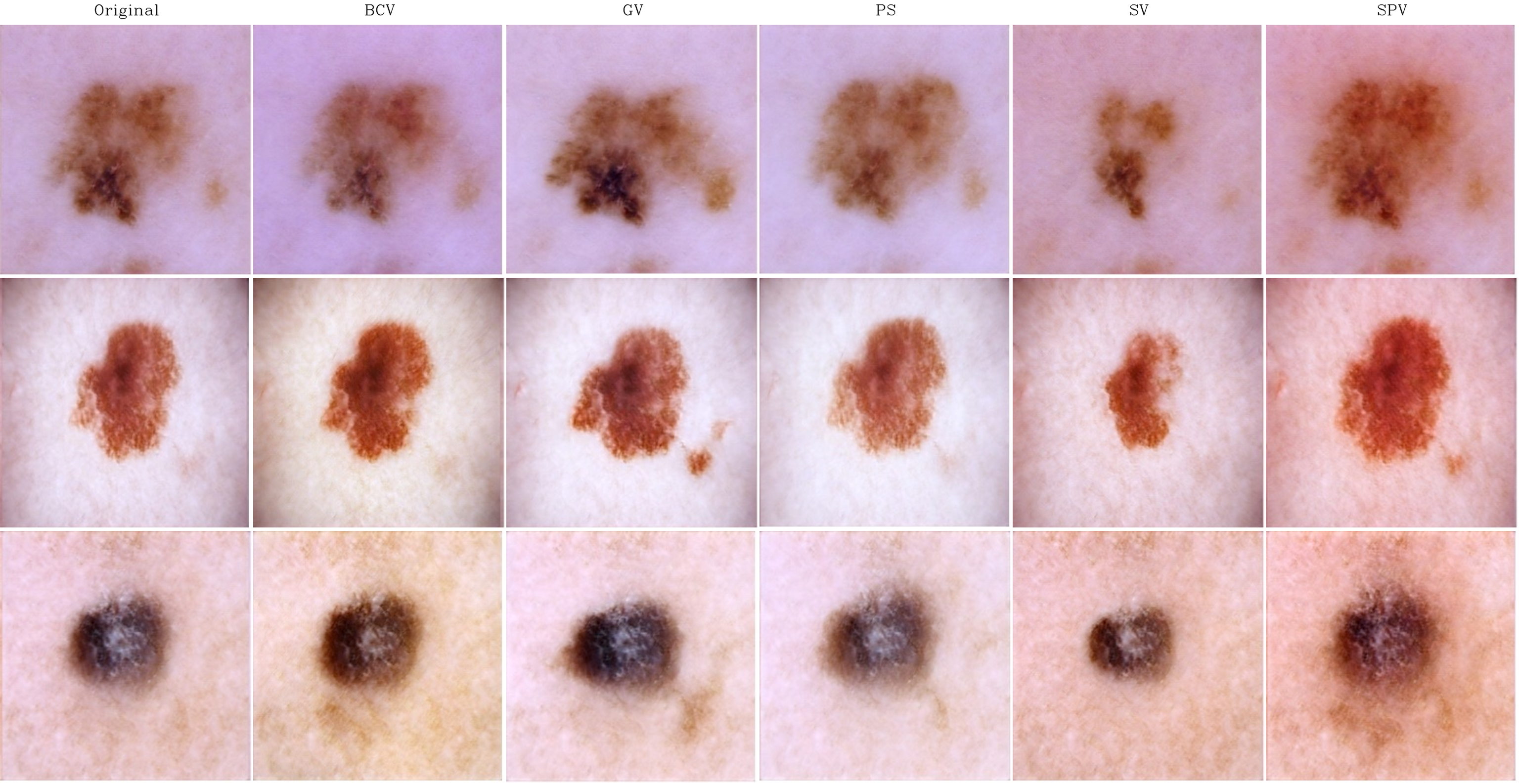}
  \caption{Original image (left) and its five transformations: BCV, GV, PS, SV, SPV (from left to right). Each transformation modifies the image in distinct ways.}
  \label{fig:app:Combined_Transformations}
\end{figure}

\setcounter{figure}{0} 

\section{Explanations of the Baseline and the Augmented Models for the class MEL}
\begin{figure}[!h]
    \centering
    \includegraphics[width=0.5\linewidth]{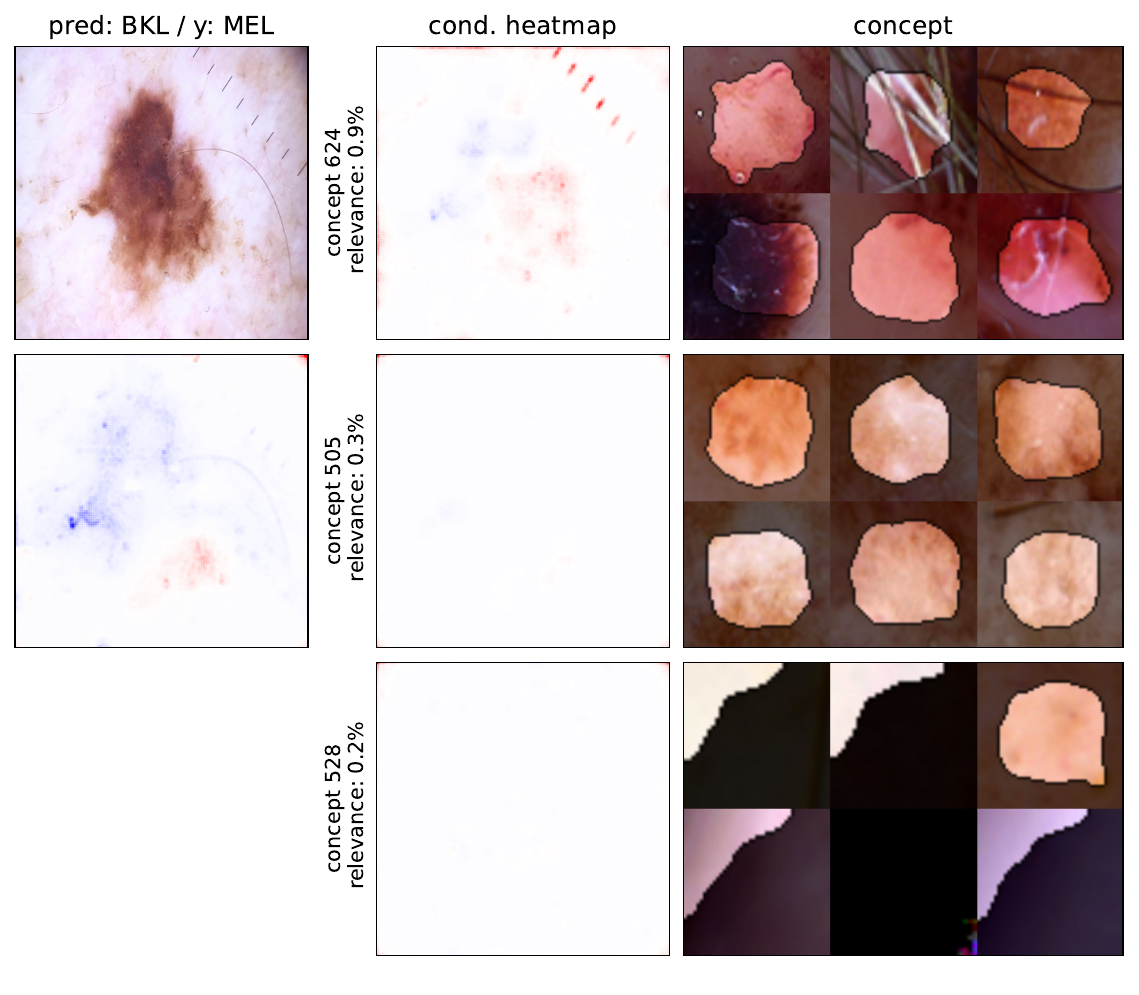}
    \caption{Explanation of the same test sample from Figure \ref{fig:crp_analysis}, for the baseline model and the ground truth class MEL. }
    \label{fig:app:crp_groundtruth}
\end{figure}

CRP requires preselecting an output neuron to explain the network decision to. Whichever output we choose, the heatmaps will show how relevant each part of the input is for this output. Figure \ref{fig:app:crp_groundtruth} shows explanations for the wrong classification class. Here we provide explanations for the ground truth class. They indicate that the baseline model is incapable of finding any supporting evidence for the MEL class, contrary to the augmented model.

\setcounter{figure}{0} 

\section{Understanding Prediction Strategies with Prototypical Concept-based Explanations}\label{appendix:pcx_section}
One of the major categorizations of Explainable AI methods is the contrast between local and global explanations. Local explanations shed light on the model behavior on a specific test sample, whereas global methods explain the model’s reasoning in general, in a holistic fashion. CRP outputs local concept conditional heatmaps, as well as global explanations of each concept. This is why it is referred to as a glocal method \cite{achtibat2023attribution}.

Similarly, Prototypical Concept Explanations (PCX) focus on a single class to provide explanations revealing the different substrategies used for the classification decisions to this class. Each substrategy is further analyzed to identify the driving concepts for decisions using this substrategy. This constitutes a global explanation. Furthermore, each local decision can be attributed to a substrategy, or identified as a novelty for the model.

Specifically, \gls{pcx} clusters latent relevances, for instance obtained with \gls{lrp}, for samples from one class (here: MEL), followed by a cluster analysis, e.g., with Gaussian Mixture Models. 
This produces clusters of samples for which the model uses similar prediction strategies. 
Each cluster can be represented with prototypical samples and viewed as distribution over concepts. These concepts can further be visualized using \gls{crp}. Figures \ref{fig:app:pcx_base} and \ref{fig:app:pcx_synth} portray global explanations for the MEL class, for the models trained on the vanilla and synthetically augmented datasets, respectively. Specifically, columns show prototypes per cluster and rows represent concepts visualized with CRP.
The values in the matrix indicate how much a concept is used by a prototype. Note that each sub-strategy can be considered as a distribution over concepts. The weight of each concept per substrategy is visualized as percentage in the matrix. While the baseline model heavily relies on distractor concepts, such as concepts 624 and 180, focusing on hair and skin markers, the model augmented with synthetic data uses clean and, to the best of our knowledge, clinically meaningful features. 

\section{Understanding Prediction Strategies with Prototypical Concept-based Explanations}
One of the major categorizations of Explainable AI methods is the contrast between local and global explanations. Local explanations shed light on the model behavior on a specific test sample, whereas global methods explain the model’s reasoning in general, in a holistic fashion. CRP outputs local concept conditional heatmaps, as well as global explanations of each concept. This is why it is referred to as a glocal method\cite{achtibat2023attribution}.

Similarly, Prototypical Concept Explanations (PCX) focus on a single class to provide explanations revealing the different substrategies used for the classification decisions to this class. Each substrategy is further analyzed to identify the driving concepts for decisions using this substrategy. This constitutes a global explanation. Furthermore, each local decision can be attributed to a substrategy, or identified as a novelty for the model.

\begin{figure}[tb]
  \centering
  \includegraphics[origin=c,width=0.85\linewidth]{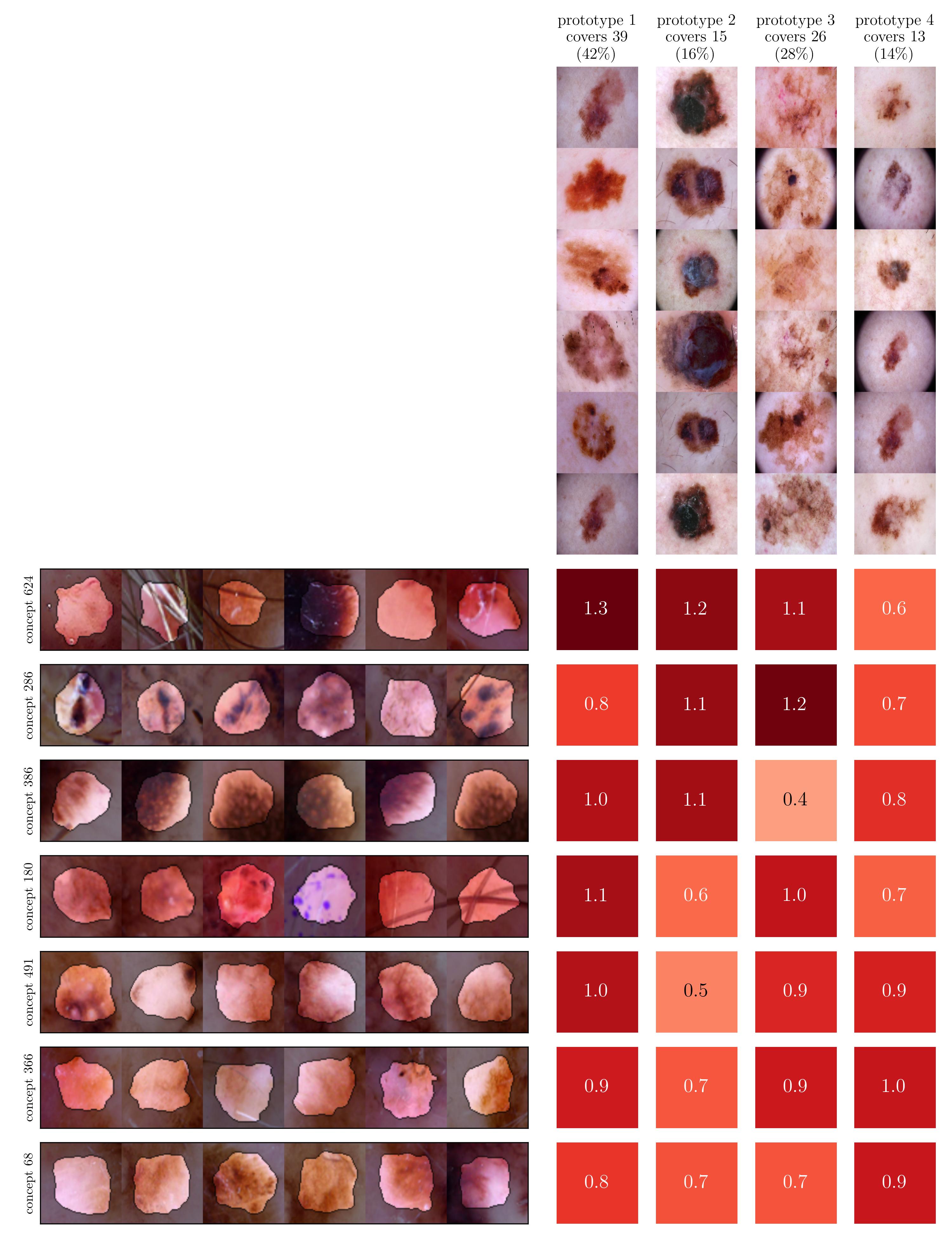}
  \caption{PCX visualization of baseline model for class MEL}
  \label{fig:app:pcx_base}
\end{figure}

\begin{figure}[tb]
  \centering
  \includegraphics[origin=c,width=0.85\linewidth]{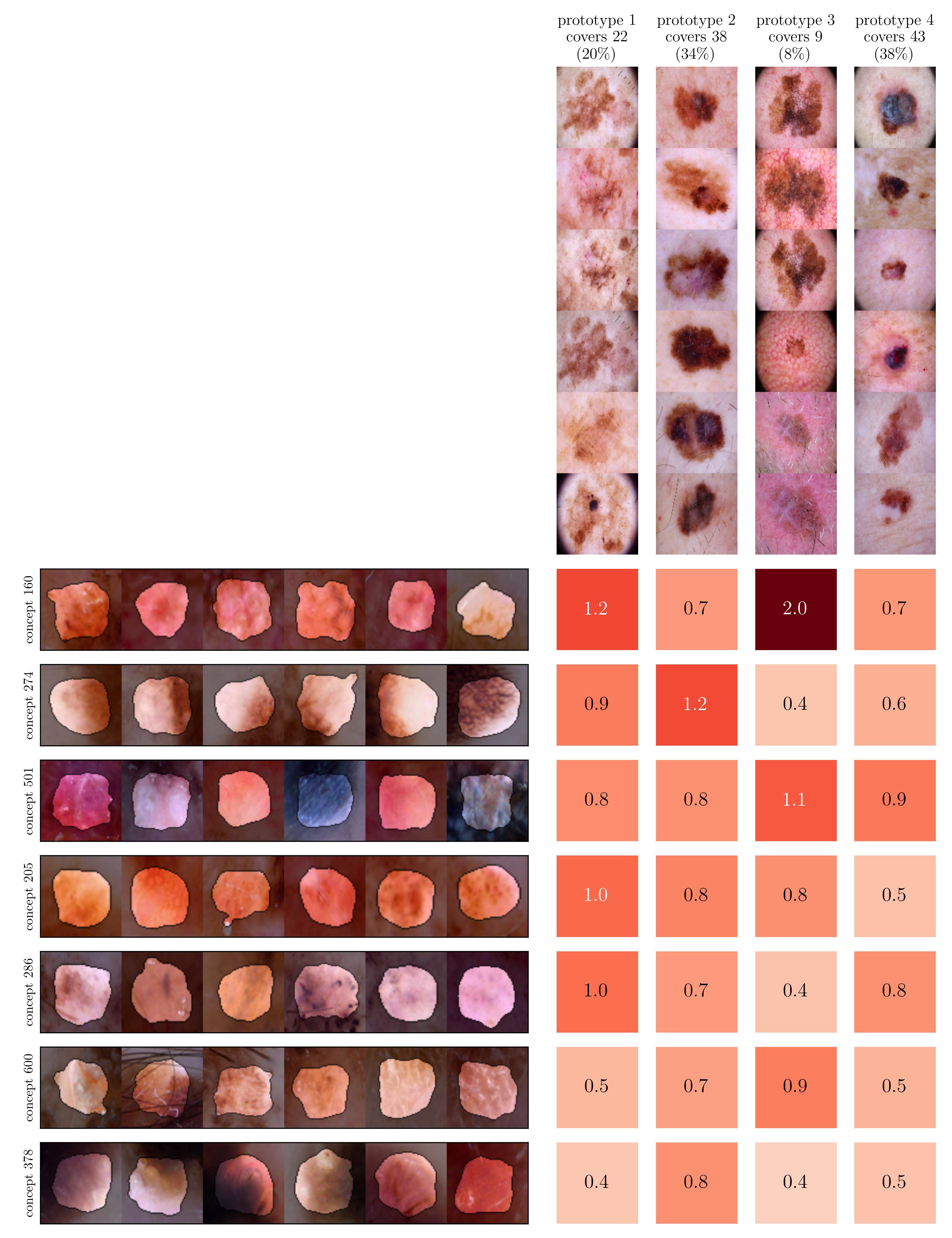}
  \caption{PCX visualization of model trained with additional (filtered) synthetic samples for class MEL}
  \label{fig:app:pcx_synth}
\end{figure}

Figures \ref{fig:app:pcx_base} and \ref{fig:app:pcx_synth} portray global explanations for the MEL class, for the models trained on the vanilla and synthetically augmented datasets, respectively. The columns in the figures correspond to different substrategies from the classification model, as discovered by a Gaussian Mixture Model trained on latent relevance scores. Subtrategies are visualized with exemplary prototypes from the training dataset. The rows correspond to different concepts on a preselected layer (here: activations \emph{after} the last transition block) and show CRP-style concept representatives. Note, that each sub-strategy can be considered as distribution over concepts. The weight of each concept per substrategy is visualized as percentage in the matrix. While the baseline model heavily relies on distractor concepts, such as concepts 624 and 180, focusing on hair and skin markers, the model augmented with synthetic data uses clean and, to the best of our knowledge, clinically meaningful features. 

\end{document}